\documentclass{article}

\usepackage{PRIMEarxiv}

\usepackage[utf8]{inputenc} 
\usepackage[T1]{fontenc}    
\usepackage{hyperref}       
\usepackage{url}            
\usepackage{booktabs}       
\usepackage{amsfonts}       
\usepackage{nicefrac}       
\usepackage{microtype}      
\usepackage{lipsum}
\usepackage{fancyhdr}       
\usepackage{graphicx}       
\usepackage{multirow}
\usepackage{cite}
\graphicspath{{media/}}     

\title{A Sequential Meta-Transfer (SMT) Learning to Combat Complexities of Physics-Informed Neural Networks: Application to Composites Autoclave Processing
}

\author{
  Milad Ramezankhani \\
  Materials and Manufacturing Research Institute \\
  The University of British Columbia \\
  Kelowna, V1V 1V7, Canada.\\
  \texttt{milad.ramezankhani@ubc.com} \\
   \And
  Abbas S. Milani \\
  Materials and Manufacturing Research Institute \\
  The University of British Columbia \\
  Kelowna, V1V 1V7, Canada.\\
  \texttt{abbas.milani@ubc.com} \\
}

\begin{document}
\maketitle

\begin{abstract}
Physics-Informed Neural Networks (PINNs) have gained popularity in solving nonlinear partial differential equations (PDEs) via integrating physical laws into the training of neural networks, making them superior in many scientific and engineering applications. However, conventional PINNs still fall short in accurately approximating the solution of complex systems with strong nonlinearity, especially in long temporal domains. Besides, since PINNs are designed to approximate a specific realization of a given PDE system, they lack the necessary generalizability to efficiently adapt to new system configurations. This entails computationally expensive re-training from scratch for any new change in the system. To address these shortfalls, in this work a novel sequential meta-transfer (SMT) learning framework is proposed, offering a unified solution for both fast training and efficient adaptation of PINNs in highly nonlinear systems with long temporal domains. Specifically, the framework decomposes PDE's time domain into smaller time segments to create "easier" PDE problems for PINNs training. Then for each time interval, a meta-learner is assigned and trained to achieve an optimal initial state for rapid adaptation to a range of related tasks. Transfer learning principles are then leveraged across time intervals to further reduce the computational cost.Through a composites autoclave processing case study, it is shown that SMT is clearly able to enhance the adaptability of PINNs while significantly reducing computational cost, by a factor of 100.
\end{abstract}


\section{Introduction}

\subsection{Physics-informed neural networks and their current drawbacks}

Recently, Physics-Informed Neural Networks (PINNs)  \cite{raissi2019physics} have gained unprecedented popularity among the research community and have proven to be highly advantageous in both scientific and industrial applications where prior knowledge of the system’s underlying physics exists  \cite{psaros2022}. There are several reasons behind their prominence. Firstly, PINN models provide a powerful and flexible framework for integrating physical laws and constraints into the architecture of neural networks. By explicitly encoding domain-specific knowledge, namely, governing equations or boundary conditions, PINNs leverage a physics-guided approach to learning the unknown solution. Another major advantage of PINN models is the potential to significantly reduce computational costs compared to traditional simulation techniques (e.g., finite element method). PINNs can leverage the efficiency of neural network computations and their adaptability to new tasks and provide fast predictions and lower the computational costs. This is in particular very valuable in process optimization tasks which require an extensive and fast exploration of large and high-dimensional design spaces  \cite{cuomo2022scientific}. 

Despite promising results in a wide range of applications, it has been shown that conventional PINNs exhibit poor performance and fail to accurately approximate the behaviour of systems with \textit{\textbf{strong non-linearity}}  \cite{mattey2022a} and \textit{\textbf{long temporal domains} } \cite{meng2020ppinn} \cite{wang2023long}.Earlier \textit{} studies have captured the training challenges of conventional PINNs when learning PDE systems with highly nonlinear time-varying characteristics and sharp transitions (i.e., stiff PDEs where the solution exhibits a significant disparity in time scales) \cite{krishnapriyan2021characterizing, hu2021when,wang2022respecting,penwardena}. Similarly, when employed to learn an intricate system with long time domains, PINNs often return poor and sub-optimum performance  \cite{wang2023long}. Various reasons can contribute to this behaviour. One explanation could be the effect of the F-principle in the neural networks  \cite{xutraining}. It has been shown that neural networks have a learning bias toward low-frequency functions and thus can fail to learn high-frequency components and highly nonlinear functions if not trained sufficiently  \cite{rahaman2019on}. Another reason can be the presence of large values in the temporal domain (e.g., simulating a five-hour-long manufacturing process) which plays a part in saturating the activation functions and impeding the proper training  \cite{wang2023long}.In addition to PINNs’ limitation in approximating  stiff PDEs and long temporal domains, their performance can also suffer from other reasons, such as poor collocation points sampling, unbalanced loss terms, and inefficient network architecture  \cite{krishnapriyan2021characterizing} \cite{wang2021understanding}.Another major drawback of PINNs is that they are designed to approximate a  specific realization of a PDE system  \cite{wang2022mosaic}, and are \textbf{\textit{not readily generalizable}}. In other words, conventional PINNs are problem-specific by nature, and they need to be retrained from scratch for each and every new system configuration (e.g., when a new set of boundary or initial conditions is introduced for the given system). 

\subsection{State-of-the-art to cope with PINNs drawbacks}

Various methods in the literature have been proposed to alleviate the above drawbacks of PINNs in highly nonlinear systems. Namely, adaptive training and sampling  \cite{nabian2021efficient} \cite{wang2022when}, sequential learning  \cite{mattey2022a} \cite{wight2020solving}, domain decomposition  \cite{wang2022mosaic} \cite{jagtap2021extended} and novel network architectures and activations functions  \cite{liang2021reproducing} \cite{buquadratic} are among the strategies that have shown promising results in addressing PINNs’ training complications.  Wang et al.  \cite{wang2021understanding} proposed a residual weighting strategy that balances the interaction between the PINN’s loss components. Levi et al.  \cite{mcclenny2020self} proposed a soft attention method which adaptively updates the weights of training points based on how difficult they are to be learned. Sequential learning is another avenue of research that has shown promise in enhancing the PINNs training for nonlinear systems. Specifically in sequential learning, the temporal domain is discretized into small training time segments, and the PINN model is trained on ``simpler" problems in a sequential manner. Mattey et al.  \cite{mattey2022a} Introduced a sequential strategy with backward compatibility (bc-PINN) in which the temporal domain is broken down into smaller segments and each segment is trained sequentially using a single network. In a lifelong learning fashion  \cite{parisi2019continual}, the network is updated at each time interval to adapt to the new domain while ensuring learned knowledge from the previous time segments is preserved. Wang et al.  \cite{wang2022respecting} further enhanced this strategy by proposing temporal weights assigned to collocation points in an attempt to respect physical causality for training PINNs. Similarly, in  \cite{wight2020solving}, a sequence-to-sequence learning approach is proposed in which instead of employing a single network to learn all the segments, each time interval is learned by an individual \textit{subnetwork}. In this approach, known as Time Marching (TM), the initial condition of each segment is determined by the latent function learned by the subnetwork in the previous time interval (see section 3.1 for details). Furthermore, the training procedure can be facilitated by implementing TL via initializing each subnetwork with the learned weights of the subnetwork trained in the preceding time segment  \cite{penwardena} \cite{ramezankhani2021making}. Since the subnetworks are trained toward learning a similar task (solving the same PDEs under slightly different initial and boundary conditions), the latent features learned by a network trained on the time interval [\( t_{n-1}, t_{n}]\), encompass useful information for efficient learning of the next interval [\( t_{n}, t_{n+1}]\), and thus the new subnetwork can leverage such information via TL. In  \cite{meng2020ppinn},the authors introduced a  similar strategy called Parallel PINN (PPINN) with the objective of reducing the \textit{computational costs} of PINN models in long temporal domains. PPINN discretizes long temporal domains into independent and short time segments and solves them in parallel using a coarse-grained solver. A by-product of this method as elaborated above is the improvement of PINN’s performance in highly nonlinear systems.

Recently, some attempts have been made to enhance PINNs’ adaptability (generalizability). These studies have harnessed the idea of knowledge transferability among relevant tasks and implemented different methods such as transfer learning (TL)  \cite{goswami2020transfer} \cite{chen2021transfer}, multi-task learning (MTL)  \cite{desai2021one} \cite{zou2023}, multi-fidelity learning  \cite{ramezankhania} \cite{meng2020a} and meta-learning \cite{bihlo2023improving,liu2022a,li2022}. The idea is to leverage the existing knowledge, such as known physics and low-fidelity labeled data, to expedite the convergence of PINNs as well as to achieve better generalization performance. TL leverages the knowledge and learned representations from a \textit{source} task toward improving the performance on a related but different task called \textit{target}. Chen et al.  \cite{chen2021transfer} constructed different sub-tasks by varying the source term and Reynolds numbers in the Navier-Stokes equation and showed that training on one sub-task and using the learned weights for initializing other sub-tasks can drastically improve the training of PINNs. Similarly, in multi-fidelity learning, a sub-class of TL, the knowledge transfer takes place between the low-fidelity and high-fidelity domains. In a case study on advanced composites manufacturing  \cite{ramezankhania}, it was shown that in the presence of low-fidelity knowledge, the training of PINNs in systems with strong nonlinearity and boundary conditions with sharp transitions becomes faster and more efficient. In MTL, by jointly learning multiple tasks (i.e., one network with multiple heads, each assigned to one task), the network can learn and benefit from the common latent representation among similar tasks and use it toward a fast and cost-effective adaptation to new unseen tasks. Desai et al.  \cite{desai2021one}proposed a n MTL framework to alleviate PINN’s high computational expense associated with training networks for different but closely related tasks. Specifically, a PINN is initially trained on a family of differential equations. Then by freezing the hidden layers and fine-tuning the final layer or training it from scratch, the network can efficiently learn the new\textit{ }tasks. Similarly, in [27], the authors introduced multi-head PINN (MHPINN) which simultaneously learns multiple tasks using a single network and then uses shared hidden layers as a basis function for fast-solving similar PDE systems. In contrast to the above methods, meta-learning, instead of transferring useful representations from other tasks, aims to learn an optimal \textit{initial state }that is suitable for fast and efficient adaptation to a family of relevant tasks [33]. Using meta-learning, Bihlo [30] explored the effect of meta-trained learnable optimizers on the convergence and error minimization of PINN models. It was shown that using a learnable optimizer parameterized by a neural network can improve the performance of conventional PINNs trained using Adam optimizers. In [31] the authors used Reptile, a scalable meta-learning algorithm to learn a proper initialization state for training the PINN model. Their work achieved a faster convergence of PINNs in a series of ODE and PDE systems such as Poisson and Burgers equations.

\subsection{Objective and novelty of the present study}

As reviewed in section 1.2, sequential learning has already shown  \cite{mattey2022a} \cite{wight2020solving}to be an effective tool to address PINNs  shortcomings in complex nonlinear systems. However, it considerably increases the computational cost as it introduces more loss terms as well as entails training multiple networks over a set of time intervals. This could result in slow training and limit the application of such strategies in solving real-world complex problems using PINNs. Besides, while some works have leveraged knowledge transferability (e.g., via utilizing TL and similar methods) towards making PINNs’ training and implementation more efficient  \cite{chen2021transfer} \cite{desai2021one} \cite{ramezankhania}, no notable research has been conducted to date on reducing the computational costs and improving the adaptability of sequential learning strategies via knowledge transferability. 

 The present work aims at developing a novel sequential meta-transfer (SMT) learning approach for more efficient, adaptable and accurate training of PINNs in highly nonlinear systems with long temporal domains. Namely, the SMT combines and leverages the well-known TL and meta-learning principles, but under a sequential learning pattern to make the training of PINNs much faster and more efficient than conventional PINNs, while ensuring high adaptability to other relevant tasks/systems (see also Figure 1). At each time segment, instead of training task-specific networks, a series of meta-learners are trained with the goal of obtaining a set of optimal initial parameters used for a fast adaptation to a range of related tasks (e.g., different boundary condition configurations).  The work also for the first time introduces an ``adaptive temporal segmentation" strategy which adaptively selects the span of the next time interval for training. For each time interval, it evaluates the performance of the subnetwork trained on the previous time segment as a measure of similarity between the two domains (e.g., if the tasks are similar, the model should perform well on both) and based on that chooses the length of the interval. This results in an efficient way of reducing the number of sequential learning steps by allocating a large step to less difficult regions in the domain and employing finer time intervals for areas with highly nonlinear behaviour.

\subsection{Application to Advanced Autoclave Processing}

In advanced composites autoclave processing, accurate prediction of the part’s temperature and degree of cure (DoC) is of paramount importance in the process design optimization tasks as it directly influences other aspects of the process outcomes, from process-induced stress and cure shrinkage strains to resin thermal degradation due to excessive exothermic reactions  \cite{fernlund2018}. In addition, the design space is often very complex and high-dimensional and thus, requires an extensive exploration by the optimization algorithms. This entails accessing an efficient surrogate model that provides accurate and fast predictions of the part’s thermal profile during the cure cycle. The complex geometry of the composite parts and the non-uniform airflow within the autoclave imposes non-uniform heat transfer coefficient (HTC) distribution throughout the part surfaces. Inconsistent resin flow during the curing of the part also introduces non-uniform fibre volume fraction which results in un-equal cure kinetics behaviour. All of this necessitates incorporating a fast and accurate surrogate model that can predict the thermal profile of the part at various locations under different process configurations. As will be demonstrated in the following sections, the proposed SMT framework is well capable of training accurate and adaptable PINN models for complex and highly nonlinear systems with long temporal domains, which can be appropriately used in autoclave composites process optimization.

The rest of the manuscript is organized as follows. Section 2 provides an overview of PINNs and meta-learning concepts. Section 3 discusses the proposed framework and the role of its components in efficient learning of PINNs in complex systems, with the application/experimentation in composites autoclave processing. Section 4 presents the experimental results and analyze the performance of the proposed framework. Section 5 concludes the paper with a summary of the contributions and highlights the potential future research direction.

\begin{figure}[!htbp]
\centering
\includegraphics[width=16cm]{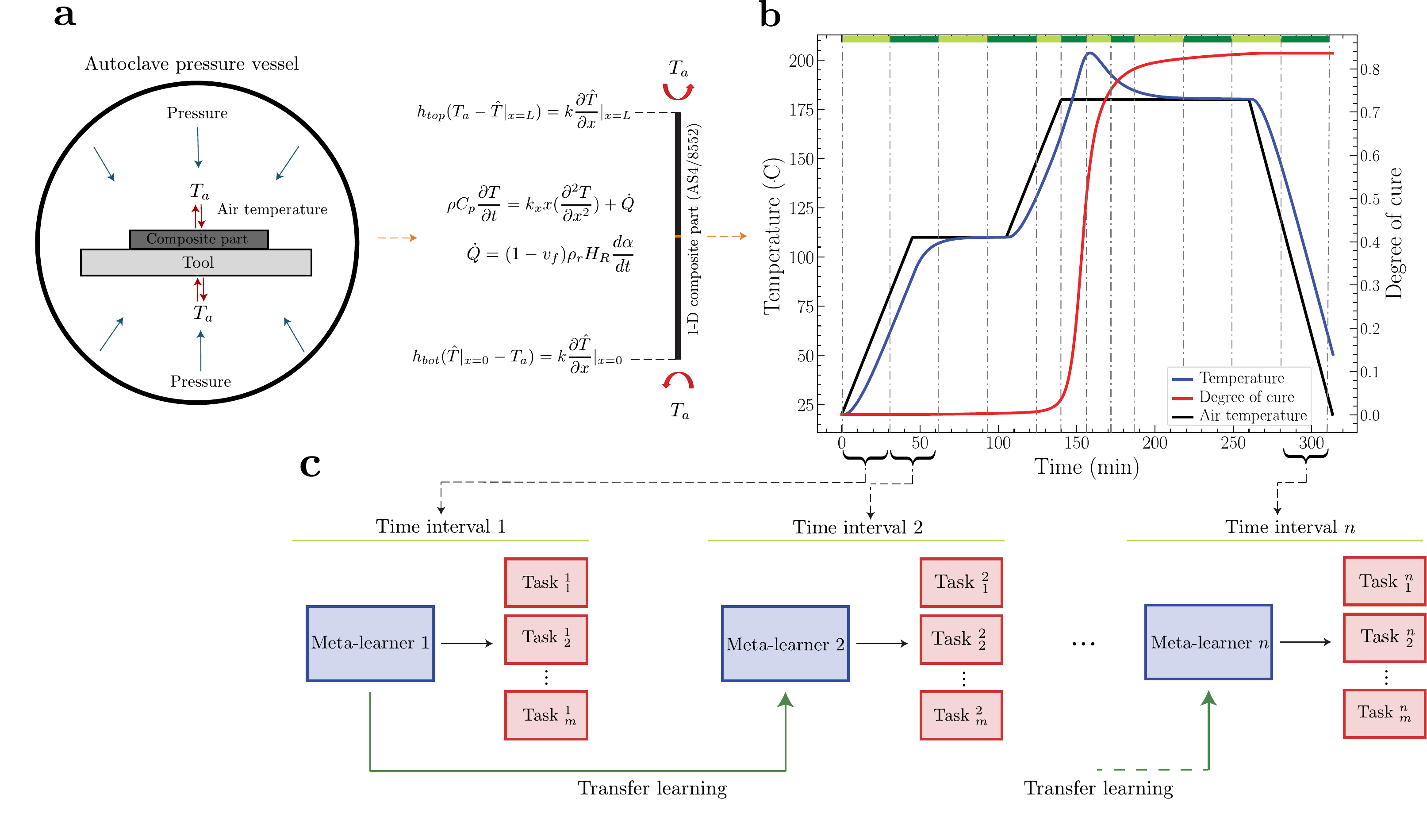}
\caption{Overview of proposed sequential meta-transfer (SMT) learning approach. (a) schematic of autoclave pressure vessel ad 1-D AS4/8552 composite part considered for this case study per section 3.4. Autoclave air temperature $T_a$ governs the part's boundary condition during the curing process. (b) Thermochemical evolution of the part's mid-section during the curing process. The black dash-dotted lines represent the time intervals used for PINNs sequential learning (Illustrated by light and dark green boxes). (c) schematic of the SMT framework based on meta-learning and TL principles. (For interpretation of the colors in the figures, the reader is referred to the web version of this article.)}
\label{fig:overview_proposed_sequential_metatransfer_smt}
\end{figure}

\section{Methodology: Basics}

\subsection{Physics-informed neural networks}

The basics of PINNs are briefly reviewed in this section, following the notations presented in  \cite{wang2022respecting}. In PINNs the solution of PDEs is inferred using neural networks and their universal approximation capabilities  \cite{raissi2019physics} \cite{wang2022respecting}. Specifically, PDEs can take the form of:

\begin{equation}
\mathbf{u}_{t}\mathcal{+N}\left[\mathbf{u}\right] = 0,       t\in\left[0,T\right],  \mathbf{x}\in \Omega ,
\end{equation}

subject to the initial and boundary conditions:

\begin{equation}
\mathbf{u}\left(0,\mathbf{x}\right) =\mathbf{g}\left(\mathbf{x}\right), \mathbf{x}\in \Omega
\end{equation}

\begin{equation}
\mathcal{B}\left[\mathbf{u}\right] = 0,  t\in\left[0,T\right],  \mathbf{x}\in \partial \Omega
\end{equation}

where \(\mathbf{u}\)\textbf{ }is the latent solution, \(\mathcal{N}\) is the PDE’s differential operator, and \(\mathcal{B}\) represents the boundary operator and it can take the form of Dirichlet, Neumann and Robin boundary conditions. \( \Omega\) and \( \partial \Omega\) denote the domain and the boundary domain, respectively. PINNs learn the solution of the PDE using a deep neural network \(\mathbf{u}^{\theta }\left(t, x\right)\) where \( \theta\) denotes the network’s parameters which are trained by minimizing the following physics-aware loss function:

\begin{equation}
\mathcal{L}\left(\mathbf{\theta }\right) = \lambda_{ic}\mathcal{L}_{ic}\left(\mathbf{\theta }\right)+\lambda_{bc}\mathcal{L}_{bc}\left(\mathbf{\theta }\right)+\lambda_{r}\mathcal{L}_{r}\left(\mathbf{\theta }\right),
\end{equation}

where 

\begin{equation}
\mathcal{L}_{ic}\left(\mathbf{\theta }\right) =\frac{1}{N_{ic}}\sum_{i = 1}^{N_{ic}}\left\vert\mathbf{u}_{\mathbf{\theta }}\left(0,\mathbf{x}_{ic}^{i}\right)-\mathbf{g}\left(\mathbf{x}_{ic}^{i}\right)\right\vert^{2},\
\end{equation}

\begin{equation}
\mathcal{L}_{bc}\left(\mathbf{\theta }\right) =\frac{1}{N_{bc}}\sum_{i = 1}^{N_{bc}}\left\vert\mathcal{B}\left[\mathbf{u}_{\mathbf{\theta }}\right]\left(t_{bc}^{i},\mathbf{x}_{bc}^{i}\right)\right\vert^{2},\
\end{equation}

\begin{equation}
\mathcal{L}_{r}\left(\mathbf{\theta }\right) =\frac{1}{N_{r}}\sum_{i = 1}^{N_{r}}\left\vert\frac{\partial\mathbf{u}_{\mathbf{\theta }}}{\partial t}\left(t_{r}^{i},\mathbf{x}_{r}^{i}\right)\mathcal{+N}\left[\mathbf{u}_{\mathbf{\theta }}\right]\left(t_{r}^{i},\mathbf{x}_{r}^{i}\right)\right\vert^{2}.\
\end{equation}

$\left\{x_{i c}^i\right\}_{i=1}^{N_{i c}}\left\{t_{b c}^i, x_{b c}^i\right\}_{i=1}^{N_{b c}}$ and $\left\{t_r^i, x_r^i\right\}_{i=1}^{N_r}$ represents the initial, boundary and collocation points, respectively, and $N$ specifies the dataset size used for training. The loss weight hyperparameter $\lambda$ determines the influence of each loss component and can be specified by the user or learned and tuned as part of PINNs training .

\subsection{Meta-learning}

Human intelligence has the advantage of being able to quickly learn new tasks by leveraging prior experiences from related tasks. That is, humans can quickly grasp new concepts or tasks by building upon their existing knowledge and experiences. Various techniques in ML have been developed in an attempt to mimic such rapid adaptation of human intelligence. Meta-learning, in particular, enables models to learn from a set of related tasks with the goal of fast generalization and adaptation to new tasks. This is as opposed to conventional ML in which models are trained on a specific task with a large dataset  \cite{rajeswaran2019meta}. With its learning-to-learn mechanism, meta-learning learns to accumulate experiences from relevant tasks and uses it toward improving the learning of new tasks using a few data points. Training a meta-learning model results in a set of \textit{optimal} parameters which can be used for initializing a base learner for fast adaptation to a new task  \cite{finnmodel}. To accomplish this task, the \textit{gradient-based }meta-learning models employ a bi-level optimization procedure. In particular, the ``inner" optimization is responsible for learning a given task (base-learner), while the ``outer" algorithm updates the base-learner in a way that improves the meta-training objective (meta-learner)  \cite{rajeswaran2019meta} \cite{hospedales2021meta}. Model-agnostic meta-learning (MAML)  \cite{finnmodel}, as the most well-known method in this category, aims to learn a set of initial parameters which require only a few gradient steps to learn a new task. In the following, a description of the meta-learning setup for a few-shot supervised learning problem is provided as per the related work described in  \cite{finnmodel} \cite{rajeswaran2019meta} \cite{sunmeta}.

For training the meta-learning model, a distribution over the training tasks \( p\left(\mathcal{T}\right)\) is considered. Each task \(\mathcal{T}_{i}\) consists of a dataset \(\mathcal{D}_{i}\) which is split into a training set \(\mathcal{D}_{i}^{tr}\) and a test set \(\mathcal{D}_{i}^{test}\). Formally, a meta-learner model (e.g., a neural network) expressed by a parameterized function \( f^{\theta }\) is considered. The training starts with the inner step of meta-learning’s bi-level optimization. Specifically, a task \(\mathcal{T}_{i}\) is drawn from \( p\left(\mathcal{T}\right)\) and the model parameters \( \theta\) are updated (e.g., via one gradient update) using the training data \(\mathcal{D}_{i}^{tr}\) and the corresponding training loss value \(\mathcal{L}_{\mathcal{T}_{i}}\left(\theta ,\mathcal{D}_{i}^{tr}\right)\):

\begin{equation}
\theta_{i}^{'}\leftarrow \theta -\alpha \nabla_{\theta}\mathcal{L}_{\mathcal{T}_{i}}\left(\theta ,\mathcal{D}_{i}^{tr}\right)
\end{equation}

where \( \alpha\) is the base learner’s step size. Then, the updated parameters \( \theta_{i}^{'}\) is used to evaluate the model performance against the test set \(\mathcal{L}_{\mathcal{T}_{i}}\left(\theta^{'},\mathcal{D}_{i}^{test}\right)\). The goal here is to leverage \(\mathcal{D}_{i}^{tr}\) toward learning task-specific parameters which minimizes the loss value of the test set. This procedure is repeated for all the tasks drawn from \( p\left(\mathcal{T}\right)\) during the training. Next, in the outer loop optimization, the calculated test losses from all the tasks used in the inner loop training phase \(\left\{\mathcal{L}_{\mathcal{T}_{i}}\left(\theta_{i}^{'},\mathcal{D}_{i}^{test}\right)\right\}_{\mathcal{T}_{i}\in p\left(\mathcal{T}\right)}\) are used to optimize the meta-learner’s parameters \( \theta\) as follows:

\begin{equation}
\theta \leftarrow \theta -\beta \nabla_{\theta }\sum_{\mathcal{T}_{i}\in p\left(\mathcal{T}\right)}^{}\mathcal{L}_{\mathcal{T}_{i}}\left(\theta_{i}^{'},\mathcal{D}_{i}^{test}\right)
\end{equation}

where \( \beta\) is the meta-learner’s step size. Once trained, then the base learner can be used as the optimal initialization state for learning new tasks with a small number of gradient steps and few-shots.

\section{Methods: Proposed Sequential Meta-Transfer (SMT) Learning and Composites Processing Case Study}

Figure 1 shows the overview of the proposed SMT framework; Figure 1.a illustrates the selected autoclave manufacturing example which will be discussed in detail in section 3.4. Concerning the SMT framework, as shown in Figure 1.b, the \textit{long} time domain is first broken down into small intervals. Leveraging the adaptive temporal segmentation, the time domain is divided into finer time intervals where the system exhibits highly nonlinear behaviour (in Figure 1.b, notice the shorter time intervals around the sharp transition of the DoC). Then, at each time segment, instead of learning task-specific network parameters, a sub-meta-learner is trained to learn a set of optimal initial parameters which enables a \textit{fast adaptation }to a range of relevant tasks (e.g., different boundary condition configurations). This is deemed a major advantage compared to the conventional PINNs for which a small change in the system settings requires training the model from scratch. PINNs trained with SMT, on the other hand, can adapt to new configurations using significantly fewer training iterations (i.e., gradient steps). The sub-meta-learners are trained in a sequential manner (Figure 1.c). Once each sub-learner is trained, it is used to initialize the meta-learner for the next time interval (via TL). Since the tasks being learned in previous time intervals are similar to the current task (i.e., the physics remains the same while the initial and boundary conditions change slightly), transferring knowledge from previously learned meta-learners can hugely facilitate the training procedure in the following time segments and hence, further increase the temporal and computational efficiency. The transferred meta-learner, on the other hand, is readily fine-tuned for the new initial and boundary conditions. 

In the sub-sections to follow, details of each comment of the proposed adaptive SMT learning framework for PINNs are described.

\subsection{Sequential learning in PINN}

This section introduces and compares the application of two sequential learning strategies, namely, TM [17] and bc-PINN [4], in composites autoclave processing. As illustrated in Figure 2, both methods revolve around the idea that decomposing the time domain into small segments facilitates the training of PINNs for highly nonlinear systems. Specifically, in TM, the time domain is divided into \( n\) time intervals. For each time interval [\( t_{i-1}, t_{i}\)], a subnetwork \( f_{i}^{\theta }\) is assigned (Figure 2.a). The training is initiated with the first time interval [\( t_{0}, t_{1}\)] using the system’s initial condition at \( t = 0\). Once \( f_{1}\) is trained, its predictions at \( t_{1}\) are used as the initial condition of the second time interval [\( t_{1}, t_{2}\)] for training the second subnetwork \( f_{2}\). This procedure is repeated until the solution of all time intervals is learned and then the individual subnetworks are combined in order to provide predictions at any point within the time domain. In contrast to TM, in bc-PINN, only \textit{one network }is used to learn the long time domain in a sequential manner.  This is done by introducing an additional loss term which ensures that the network retains the learned knowledge from previous time segments. Specifically, a new loss term is added to the PINN’s loss function (Eq. (4)) that satisfies the solution for all the previous time intervals:

\begin{equation}
\mathcal{L}\left(\mathbf{\theta }\right) = \lambda_{ic}\mathcal{L}_{ic}\left(\mathbf{\theta }\right)+\lambda_{bc}\mathcal{L}_{bc}\left(\mathbf{\theta }\right)+\lambda_{r}\mathcal{L}_{r}\left(\mathbf{\theta }\right)+\lambda_{LL}\mathcal{L}_{LL}\left(\mathbf{\theta }\right).
\end{equation}

Here, \( \mathcal{L}_{LL} \) represents the loss accounting for the departure of the network from the already learned solution in the previous time intervals. After training at each time step, the learned weights are used to make predictions at some random data points from the same time intervals and the predictions are stored as “true” labels to be used for calculating \( \mathcal{L}_{LL} \) when training the subsequent time intervals. As demonstrated in Figure 2.b and similar to TM, the time domain is initially divided into n time segments. At each time step, the network h $\theta$ is trained/fine-tuned to learn the underlying solution via the initial condition, boundary condition and the PDE loss components, while ensuring knowledge retention from previous time steps using the added loss term. This means that for learning the time interval [\(t_{n-1}, t_n\)], bcPINN requires accessing the network's predictions in all previously learned time intervals [\(0, t_{n-1}\)].This is as opposed to TM in which only the learned latent solution in the previous time interval [\(t_{n-2}, t_{n-1}\)]

\begin{figure}[!htbp]
\centering
\includegraphics[width=16cm]{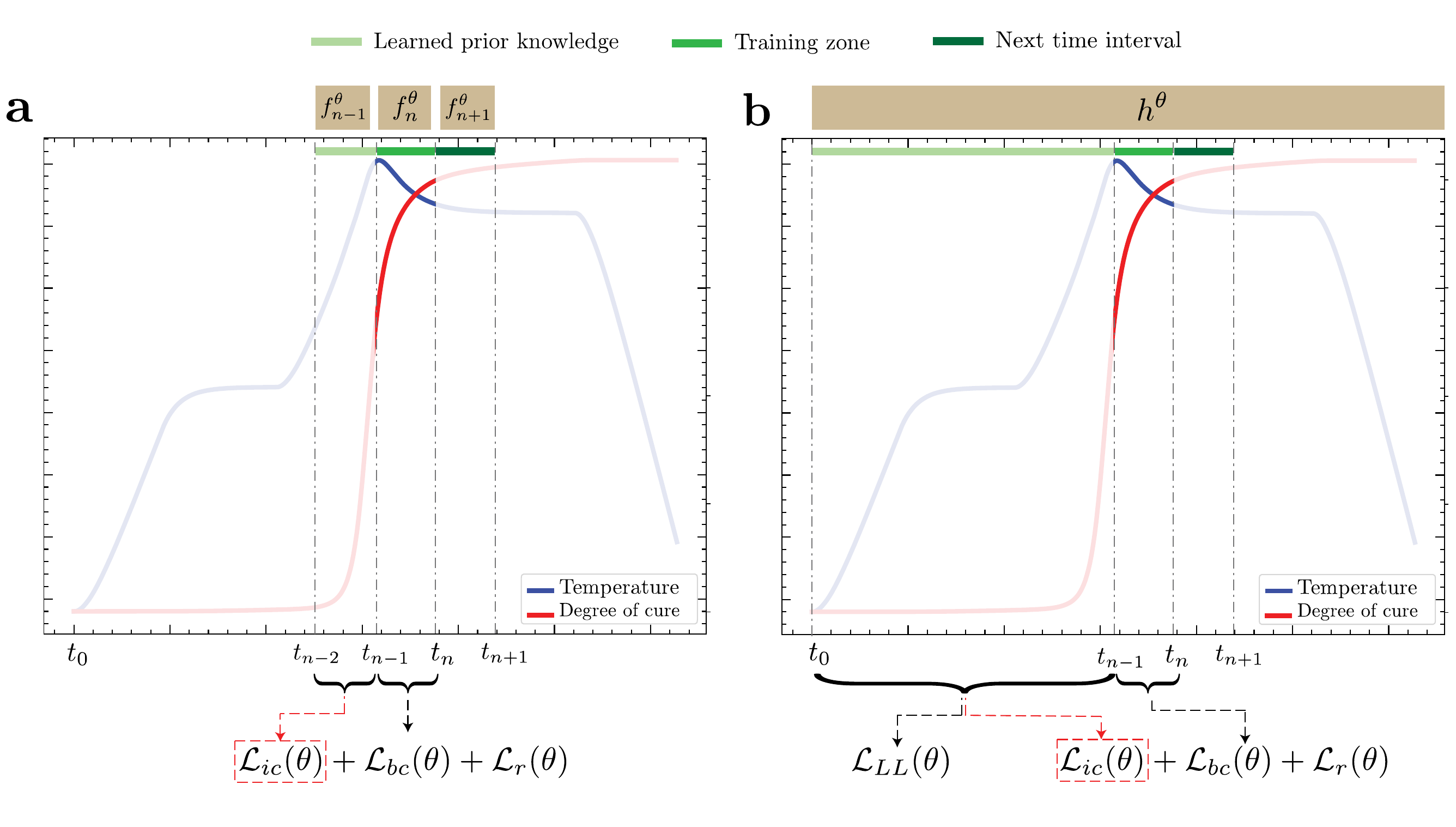}
\caption{Schematic of TM (a) and bcPINN (b) sequential learning approaches. Both methods use the ``learned prior knowledge" from previous time intervals as part of learning the ``training zone". Specifically, the initial condition at \( t_{n}\) is determined by the learned latent solution in the previous time intervals (indicated by red dashed line). In TM, each time interval is learned by an individual subnetwork \( f_{i}^{\theta }\), whereas in bcPINN one network \( h^{\theta }\) is employed to estimate the solution in a sequential manner. The additional loss term \(\mathcal{L}_{LL}\left(\theta\right)\) in bcPINN is responsible for maintaining the prior learned knowledge across all the previous time intervals.}
\label{fig:b_and_similar_tm_time}
\end{figure}

\subsection{Meta-transfer learning}

Sequential learning methods discussed in section 3.1 focuses on learning a specific realization of a PDE system. For instance, the nonlinear thermochemical behavior of a composite part in an autoclave with a specific set of configurations (i.e., initial and boundary conditions, and material properties) can be captured using the TM approach by training subnetworks on multiple time intervals. However, the proposed SMT method aims to train a model that enables efficient and rapid adaptation of PINNs to a \textit{distribution} over tasks. To accomplish this, the subnetworks in TM approach are replaced with \textit{meta-learners }and they are trained using a support set, consisting of a set of tasks defined for training purposes. Specifically in this study, various tasks are defined by modifying the boundary condition parameters (see section 3.4 for details.) 

One particular limitation of meta-learning is the method’s poor performance against out-of-distribution tasks. This becomes more evident when these tasks introduce a drastic domain shift  \cite{sunmeta}. This is especially true in the sequential learning of PINNs where transitioning from one time interval to another can result in significant discrepancies in initial/boundary conditions and the system’s behavior. In such a scenario, one might need to train the meta-learners from scratch for each time interval in order to ensure an efficient adaptation to new tasks. However, this can be infeasible in PINNs’ sequential learning setting as training the meta-learners from scratch for multiple time intervals is exhaustively demanding and time-consuming. One solution would be to \textit{transfer} the learned knowledge from the trained meta-learner at hand to the training of the subsequent meta-learners in the following time intervals. Inspired by the work in  \cite{sunmeta}, here, a \textit{meta-transfer} learning strategy is employed to make the training of the proposed sequential framework more efficient (Figure 1.c). In particular, the training begins with training the first meta-learner (ML1) using the initial and boundary conditions of the first time interval \(\left[0, t_{1}\right]\). Once trained, ML1’s learned weights are used to initialize the second meta-learner (ML2) trained for the second time interval \(\left[ t_{1}, t_{2}\right]\). This process is repeated until the meta-learners associated with all time intervals are trained.

\subsection{Adaptive temporal segmentation}

Sequential learning methods have proven to be effective in accurately learning regions with rapid shifts and kinks in the boundary condition or highly nonlinear behaviour in the latent solution function (e.g., temperature and DoC in composites processing). It also has been shown that employing smaller time segment sizes can significantly improve the PINN’s performance. However, this comes with the trade-off of more computational costs as it entails training more subnetworks. Thus, one would like to avoid making the time interval \textit{too }small. To ensure that the segment length is appropriately proportionate to the complexity of the system’s behaviour and to avoid unnecessary computational expenses, the following adaptive segmentation strategy is implemented. Initially, the temporal domain is divided into \( n\) segments with equal lengths and the training begins with the first interval (Figure 1.b). Once the first subnetwork is trained and a desirable training/test loss is achieved, the optimized weights are transferred to initialize the next subnetwork. Then, before initiating the training, the training loss of the new subnetwork (with initialized weights) on the second time interval is calculated. The loss value can be viewed as a representative of the level of discrepancy between the source and target tasks as well as the difficulty level of the target task (e.g., in terms of nonlinearities). The loss value on the new task is then compared against the training loss of the previous task (source) and if the difference exceeds a user-defined threshold \( \epsilon\), the new time segment is halved and the loss value for the new time segment (now with half-length of the original size) is calculated and compared. This step is repeated until an acceptable initial loss is attained and then the training process begins. The benefit of the adaptive segmentation approach is two-fold. First, reducing the length of the time interval makes the training of ``stiff" and highly nonlinear systems easier. Second, from the TL point of view, a large loss value on the target task using the weights of the source network signals a considerable discrepancy between the source and target tasks (here, neighbouring time intervals), and thus, yields a poor TL performance. By shortening the time interval, the farther points from the source domain are removed which results in a target domain with an input space closer to the source domain. Above all, this strategy ensures that more computational capacities are allocated only to ``difficult" regions.

\subsection{Experimentation: Autoclave processing case study}

Autoclave processing is a widely employed method in the manufacturing of advanced composite structures. During this process, the composite part undergoes a pre-defined temperature and pressure cycle, known as the ``cure cycle", with multiple heat ramps and isothermal stages enforced by the autoclave’s air temperature  \cite{fernlund2018}. The objective is to cure the resin matrix in a way that an optimal resin-fibre distribution is obtained and the void/defect occurrence is minimized. The quality of the manufactured part highly depends on the process configurations as well as the properties of the raw materials employed. Temperature and DoC (indicative of resin’s chemical advancement) are two of the key \textit{state variables} in the manufacture of composite materials, influencing not only the thermochemical behaviour of the part but also the resin flow, the part’s residual stress propagation and deformation  \cite{johnston1997an}.Due to the complex nature of the curing process, the part’s temperature and  DoC exhibit a nonlinear evolution with rapid shifts (Figure 1.b).

The thermochemical behaviour of composites during the curing process is governed by an anisotropic heat conduction equation equipped with an internal heat generation term \(\dot{Q}\) representing the exothermic curing reaction of the resin matrix  \cite{johnston1997an}:

\begin{equation}
\frac{\partial }{\partial t}\left(\rho C_{p}T\right) =\frac{\partial }{\partial x}\left(k_{xx}\frac{\partial T}{\partial x}\right)+\frac{\partial }{\partial y}\left(k_{yy}\frac{\partial T}{\partial y}\right)+\frac{\partial }{\partial z}\left(k_{zz}\frac{\partial T}{\partial z}\right)+\dot{Q}
\end{equation}

where \( \rho\) denotes the part’s density, \( C_{p}\) is the specific heat capacity and \( k_{ii}\) represent anisotropic thermal conductivity coefficients. They can be calculated using local resin and fiber properties as well as fiber volume fraction. The heat generation term \(\dot{Q}\) in (11) can be expressed as:

\begin{equation}
\dot{Q} =\frac{d\alpha }{dt}\left(1-v_{f}\right)\rho_{r}H_{R}
\end{equation}

where \( \alpha\) represents the resin’s DoC, \( v_{f}\) is the fiber volume fraction, \( \rho_{r}\) is the resin density and \( H_{R}\) is the resin heat of reaction, a measure of the total amount of heat produced during a complete resin curing cycle. \(\frac{d\alpha }{dt}\) is the cure reaction rate and it is governed by the cure kinetics of the resin system. For a one-dimensional heat transfer system, Eq. (11) can be reduced to:

\begin{equation}
\rho  C_{p}\frac{\partial T}{\partial t} = k_{xx}\frac{\partial^{2}T}{\partial x^{2}}+\left(1-v_{f}\right)\rho_{r}H_{R}\frac{d\alpha }{dt}.
\end{equation}

For the curing process of a composite system with thermoset resin systems, the cure rate \(\frac{d\alpha }{dt}\) is governed by the resin’s cure kinetics and often is described as an ordinary differential equation. Specifically for 8552 epoxy (the resin system used in this paper), the cure kinetics have been already developed in previous studies  \cite{johnston1997an} and can be expressed as:

\begin{equation}
\frac{d\alpha }{dt} =\frac{K\alpha^{m}(1-\alpha )^{n}}{1+e^{C\left\{ \alpha -\left(\alpha_{C0}+\alpha_{CT}T\right)\right\} }} \\ 
, K = Ae^{-\frac{\Delta E}{RT}}
\end{equation}

where \( \Delta E\) is the activation energy, \( R\) is the gas constant and \( \alpha_{C0}\), \( \alpha_{CT}\), \( m\), \( n\) and \( A\) are constants determined by experiments. Table 1 summarized the values of the parameters used in the cure kinetics equations in this study.

\begin{table}
 \caption{Summary of parameters used in heat transfer and cure kinetics governing equations}
  \centering
  \begin{tabular}{lll}
    \toprule
    Parameter   & Description   & Value         \\
    \midrule
    \( \Delta E\) & Activation energy  & 66.5 (kJ/gmol)     \\
    \( R\)     & Gas constant & 8.314     \\
    \( A\)     & Pre-exponential cure rate coefficient      & $ 1.53 \times 10^5 $ (1/s)  \\
    \( m\)     & First exponential constant      & 0.813  \\
    \( n\)     & Second exponential constant       & 2.74  \\
    \( C\)     & Diffusion constant       & 43.1  \\
    \( \alpha_{C0}\)     & Critical degree of cure at $T=0 $ K       & -1.684  \\
    \( \alpha_{CT}\)     & Critical resin degree of cure constant       & $5.475 \times 10^{-3} $ (1/K)  \\

    \bottomrule
  \end{tabular}
  \label{tab:table}
\end{table}

The initial conditions of the coupled system described above can be specified as:

\begin{equation}
T\left.\right\vert_{t = 0} = T_{0}\left(x\right)
\end{equation}

\begin{equation}
\alpha\left.\right\vert_{t = 0} = \alpha_{0}\left(x\right)
\end{equation}

\( T_{0}\) denotes the part’s initial temperature and is often considered uniform throughout the part. This study assumes 20 $^{\circ}$C for the part’s temperature at the beginning of the curing process. \( \alpha_{0}\) is the initial DoC of the resin system and for an uncured part, it is assumed to be zero or a small value (in this study, a value of 0.001 is assumed.)

The boundary conditions can also be specified by the autoclave air temperature \( T_{a}\left(t\right)\) prescribed by the cure cycle recipe. Specifically, Robin boundary conditions can be defined to incorporate the convective heat transfer between the composite part and the autoclave air  \cite{niaki2021physics}:

\begin{equation}
h_{t}\left(T_{a}(t)-T\left.\right\vert_{x = L}\right) = k_{xx}\frac{\partial T}{\partial x}\left.\right\vert_{x = L}
\end{equation}

\begin{equation}
h_{b}\left(T\left.\right\vert_{x = 0}-T_{a}(t)\right) = k_{xx}\frac{\partial T}{\partial x}\left.\right\vert_{x = 0}
\end{equation}

where \( h_{t}\) and \( h_{b}\) refers to the top and bottom HTC values, respectively. It was shown that the value of HTC within the autoclave is a strong function of the temperature and pressure of the autoclave air  \cite{valletan}. The presence of multiple parts and tools with various sizes and complex geometries in an autoclave introduces complex airflow patterns, resulting in considerable local variations in the air temperature, and subsequently different HTC values. This makes the already complex thermochemical analysis of a composite part more intricate as it necessitates separate evaluations of the part’s thermal profile of the part at various locations with different HTC values. 

 For training the PINN on the curing process of a 1D composite part, the following loss function is employed:

\begin{equation}
\mathcal{L}\left(\mathbf{\theta }\right) = \lambda_{ic_{T}}\mathcal{L}_{ic_{T}}\left(\mathbf{\theta }\right)+\lambda_{ic_{\alpha }}\mathcal{L}_{ic_{\alpha }}\left(\mathbf{\theta }\right)+\lambda_{bc_{t}}\mathcal{L}_{bc_{t}}\left(\mathbf{\theta }\right)+\lambda_{bc_{b}}\mathcal{L}_{bc_{b}}\left(\mathbf{\theta }\right)+\lambda_{r_{T}}\mathcal{L}_{r_{T}}\left(\mathbf{\theta }\right)+\lambda_{r_{\alpha }}\mathcal{L}_{r_{\alpha }}\left(\mathbf{\theta }\right),
\end{equation}

where

\begin{equation}
\mathcal{L}_{r_{T}}\left(\mathbf{\theta }\right) =\frac{1}{N_{r}}\sum_{i = 1}^{N_{r}}\left\vert \rho  C_{p}\frac{\partial T}{\partial t}\left(t_{r}^{i},\mathbf{x}_{r}^{i}\right)-k_{xx}\frac{\partial^{2}T}{\partial x^{2}}\left(t_{r}^{i},\mathbf{x}_{r}^{i}\right)-\left(1-v_{f}\right)\rho_{r}H_{R}\frac{d\alpha }{dt}\left(t_{r}^{i},\mathbf{x}_{r}^{i}\right)\right\vert^{2}
\end{equation}

\begin{equation}
\mathcal{L}_{r_{\alpha }}\left(\mathbf{\theta }\right) =\frac{1}{N_{r}}\sum_{i = 1}^{N_{r}}\left\vert\frac{d\alpha }{dt}\left(t_{r}^{i},\mathbf{x}_{r}^{i}\right)-\frac{K\alpha^{m}(1-\alpha )^{n}}{1+e^{C\left\{ \alpha -\left(\alpha_{C0}+\alpha_{CT}T\right)\right\} }}\left(t_{r}^{i},\mathbf{x}_{r}^{i}\right)\right\vert^{2}
\end{equation}

\begin{equation}
\mathcal{L}_{bc_{t}}\left(\mathbf{\theta }\right) =\frac{1}{N_{bc_{t}}}\sum_{i = 1}^{N_{bc_{t}}}\left\vert h_{t}\left(T_{a}(t_{bc_{t}}^{i})-T\left(t_{bc_{t}}^{i},\mathbf{x}_{bc_{t}}^{i}\right)\right)-k_{xx}\frac{\partial T}{\partial x}\left(t_{bc_{t}}^{i},\mathbf{x}_{bc_{t}}^{i}\right)\right\vert^{2}
\end{equation}

\begin{equation}
\mathcal{L}_{bc_{b}}\left(\mathbf{\theta }\right) =\frac{1}{N_{bc_{b}}}\sum_{i = 1}^{N_{bc_{b}}}\left\vert h_{b}\left(T\left(t_{bc_{b}}^{i},\mathbf{x}_{bc_{b}}^{i}\right)-T_{a}(t_{bc_{b}}^{i})\right)-k_{xx}\frac{\partial T}{\partial x}\left(t_{bc_{b}}^{i},\mathbf{x}_{bc_{b}}^{i}\right)\right\vert^{2}
\end{equation}

Subscripts \( T\), \( \alpha\), \( t\) and \( b\) refer to temperature, DoC, top and bottom sides loss components.

\section{Results and Discussion}

To evaluate the prediction performance of PINNs in the above composite autoclave processing case study, a complex cure recipe with two isothermal holds is considered (Figure 1.b). The cure cycle involves two heat ramps with equal heat rates of 2 $^{\circ}$C/min and two isothermal stages at 110 $^{\circ}$C and 180 $^{\circ}$C. A 3cm AS4/8552 prepreg is considered as the raw material for this case study. A fully-connected architecture with 5 hidden layers and 64 neurons per layer is employed for all networks. Hyperbolic tangent activation function is used in all hidden layers. The location on the 1D composite part \( x\) and time \( t\) comprise the networks’ input space and their output layer is equipped with two neurons dedicated to the prediction of the part’s temperature \( T\) and DoC \( \alpha\). For the remainder of the paper, the same network architecture is used for all case studies unless mentioned otherwise. All networks are trained using Adam optimizer with the default hyperparameters  \cite{kingma2014a}. An initial learning rate of 1$\times$10\textsuperscript{-5} with an exponential decay schedule with a decay rate of 0.9 per 5000 steps is utilized. For all networks, the weight hyperparameters \( \lambda\) in Eq. (19) are set to 1 except for initial condition terms (\( \lambda_{ic_{T}}\) and \( \lambda_{ic_{\alpha }}\)) which are set to 100. For training SMT’s meta-learners, the outer loop uses the same Adam optimizer specifications and learning rate, while the inner loop employs a one-step gradient descent with a learning rate of 1$\times$10\textsuperscript{-5}. The JAX implementation of the SMT framework will be made available at \url{https://github.com/miladramzy/SequentialMetaTransferPINNs}.

\subsection{Training PINNs with sequential learning strategies}

Three training scenarios are considered. First, the original formulation of PINN  \cite{raissi2019physics} is used to establish a baseline performance. Next, two sequential learning approaches, namely, TM and bcPINN, are implemented to address the shortfalls of conventional PINN. For both methods, the temporal domain is initially divided into 10 time segments and the segment lengths and counts are adaptively updated using the method described in section 3.3. This results in 12 intervals with the two additional intervals occurs around the sharp transition of DoC (Figure 1.b). The training specifications of the models along with their predictive performance are summarized in Table 2. The models’ predictions of the temperature and DoC of the part’s mid-point are demonstrated in Figure 3 and Figure 4. The FE simulation results for the same curing process are also overlayed for comparison. The conventional PINN exhibits a very poor prediction performance as it fails to capture the governing equations, especially the drastic shift of the DoC in the middle of the cure cycle. While bcPINN outperforms the conventional PINN and results in a close alignment with the FE simulation, it yields poor predictions at a few regions throughout the curing process. This is evident right after the kinks in the cure cycle and in the proximity of the exotherm (where the part’s temperature reaches its maximum value). These are associated with regions where learning the system’s solution is the most difficult  \cite{zobeiry2021a}.Such sudden shifts and nonlinearities  introduce significant discrepancies between the distributions of the nearby time intervals. For bcPINN which uses one network to learn all time intervals one after another, such drastic changes can cause the model to forget the knowledge from the past intervals. This is due to the fact that once the model encounters an interval with a strong nonlinearity (hence, a significant distribution shift in comparison to previous time intervals/tasks), it generates \textit{large error gradients }which destructively update the already trained weights. It was also observed that for time segments with high nonlinearities (e.g., sharp DoC transition), the model fails to remember the past learnings. In other words, the model prioritizes minimizing the ``stiff" PDE/ODE and initial and boundary losses by allowing more room for \(\mathcal{L}_{LL}\) (i.e., the loss associated with the learnings of past time intervals). This trade-off can lead to the model’s poor performance on previous time segments.

TM, however, produces the dominant performance among the studied approaches and results in the lowest prediction loss (Table 2). It can be observed that by learning complex sub-domains in a sequential manner via temporal domain decomposition, one can avoid conventional PINN’s pitfalls and achieve very accurate predictions. Additionally, in contrast to bcPINN, TM enjoys an easier optimization scheme with fewer loss components, which hugely improves its training time and accuracy. TM is advantageous since it does not require concurrently maintaining high accuracy in the previous intervals while solving PDEs in difficult regions. It was observed that employing separate subnetworks throughout the long time domains can result in better generalization performance. Thus, for the remainder of this paper, TM is used to sequentially train the PINN models as part of the SMT framework.

text 
text

\begin{table}[h]
\centering
\caption{Training specification and generalization performance of PINN, TM, and bcPINN on composite part’s temperature and DoC predictions.}
\begin{tabular}{cp{1cm}p{1.2cm}p{1.5cm}p{1cm}p{1cm}p{1.7cm}p{1.5cm}p{1.7cm}p{1.7cm}}
\hline 
\multirow{2}{*}{ Model} & \multirow{2}{1cm}{Time intervals count} & \multirow{2}{1.2cm}{Training time (epoch/s)}  & \multicolumn{3}{c}{Count per time interval} & \multicolumn{2}{c}{Temperature} & \multicolumn{2}{c}{Degree of cure} \\
\cmidrule(rl){4-6} \cmidrule(rl){7-8} \cmidrule(rl){9-10}
 &  &  & Collocation points & BC points & IC points & Relative $\mathcal{L}^2$ error & Maximum prediction error $\left({ }^{\circ} \mathrm{C}\right)$ & Relative $\mathcal{L}^2$ error & Maximum prediction error $\left({ }^{\circ} \mathrm{C}\right)$ \\
\hline
PINN [1] & 1 & 25 & 25000 & 5000 & 2000 & $3.8 \times 10^{-2}$ & 43.80 & $9.98 \times 10^{-1}$ & $8.3 \times 10^{-1}$ \\
TM $[17]$ & 12 & 95 & 2000 & 400 & 200 & \boldmath{$4.3 \times 10^{-4}$} & \textbf{1.705} & \boldmath{$2.4 \times 10^{-3}$} & \boldmath{$9.1 \times 10^{-3}$} \\
bcPINN [4] & 12 & 81 & 2000 & 400 & 200 & $5.2 \times 10^{-3}$ & 11.46 & $1.5 \times 10^{-2}$ & $5.4 \times 10^{-2}$ \\
\hline 
\end{tabular}
\end{table}

\begin{figure}[!htbp]
\centering
\includegraphics[width=16cm]{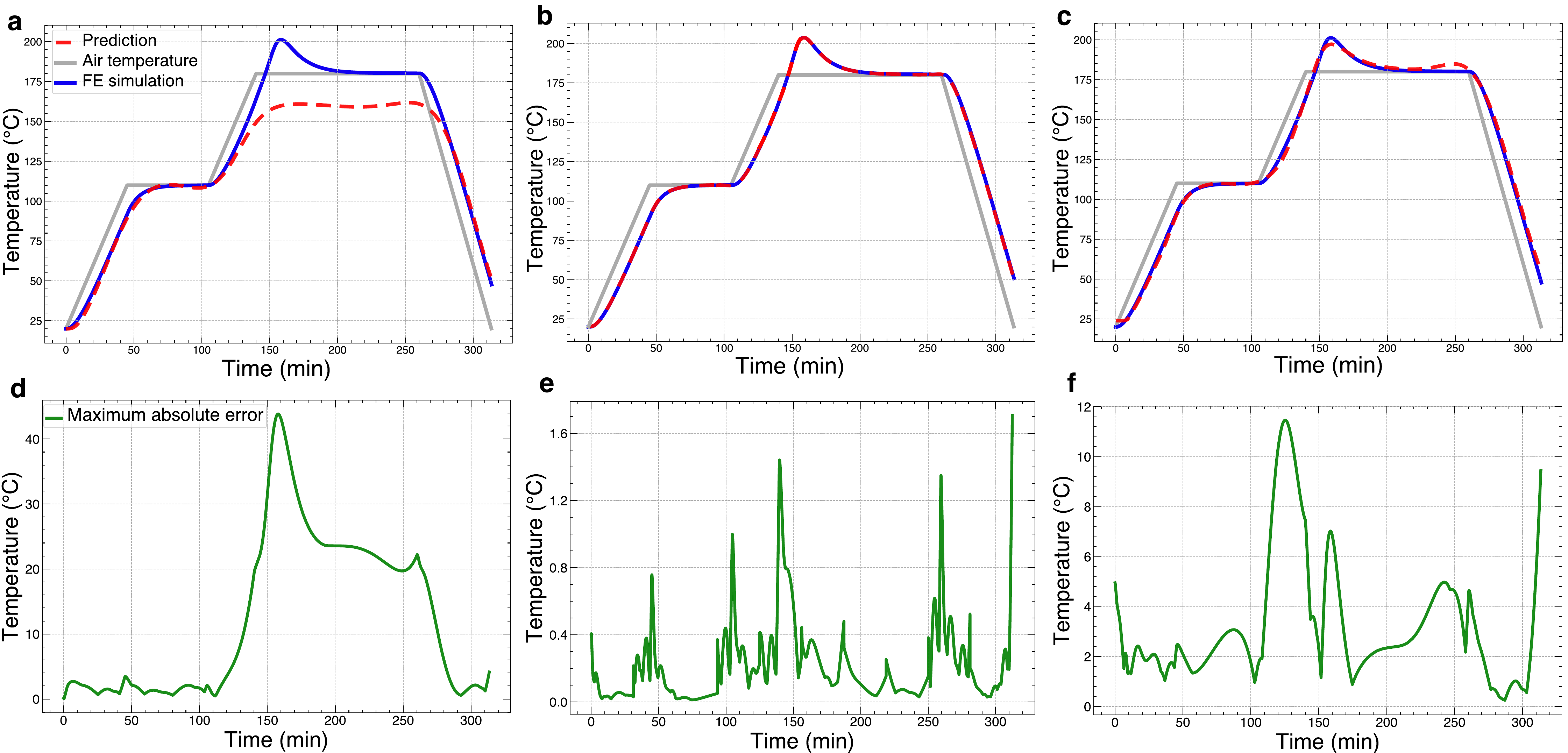}
\caption{Temperature prediction and maximum absolute error of conventual PINN (a,d), TM (b,e), and bcPINN (c,f) for the midpoint of the composite part cured in a two-hold cure cycle.}
\label{fig:temperature_prediction_and_maximum_absolute}
\end{figure}

\begin{figure}[!htbp]
\centering
\includegraphics[width=16cm]{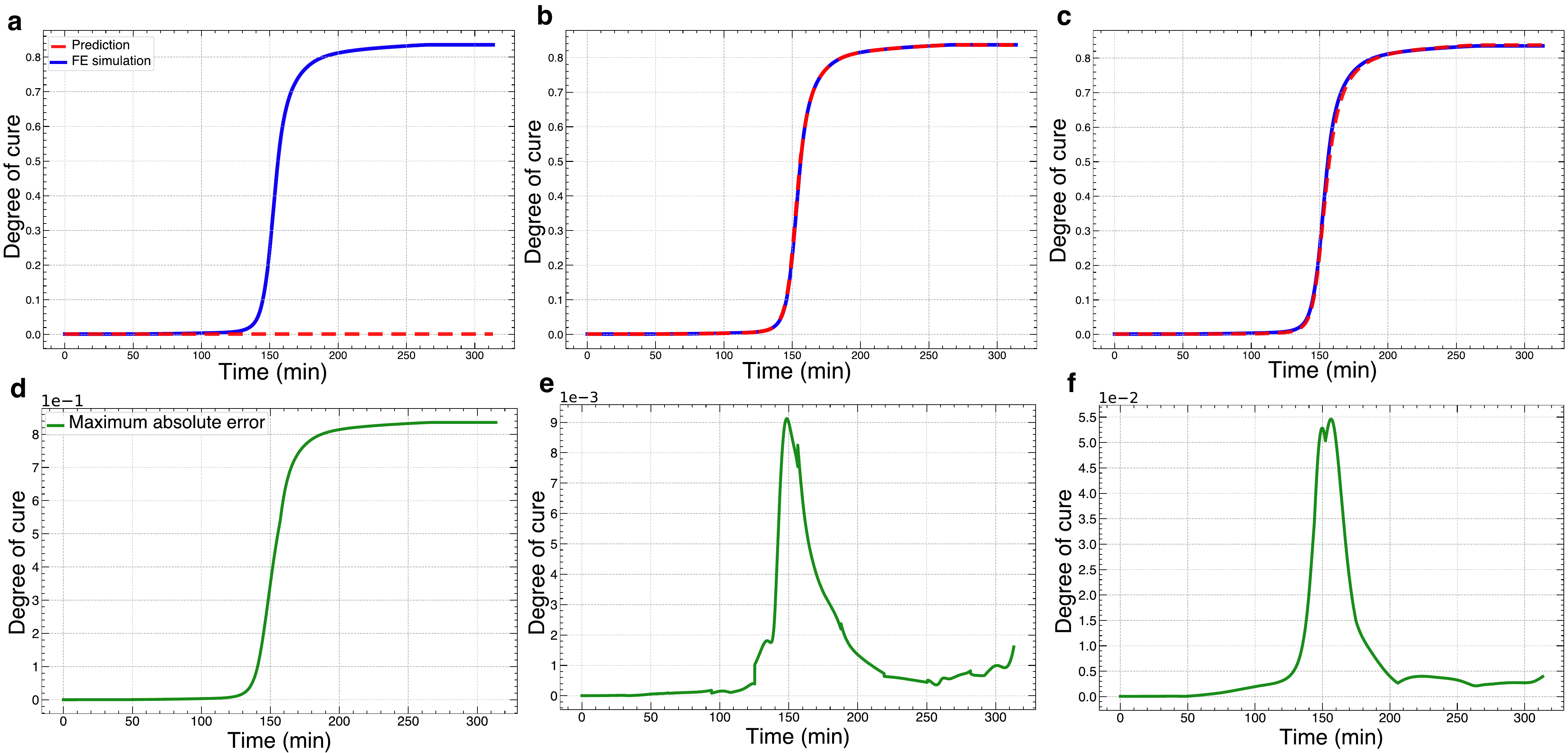}
\caption{DoC prediction and maximum absolute error of conventual PINN (a,d), TM (b,e), and bcPINN (c,f) for the mid-point of the composite part cured in a two-hold cure cycle.}
\label{fig:doc_prediction_and_maximum_absolute}
\end{figure}

\section{Assessing fast task adaptation of sequential meta-transfer learning}

As was shown in Section 4.1, TM can achieve superior generalization performance in learning the thermochemical behaviour of composite materials during the curing process in an autoclave. However, it involves multiple stages of training PINN sub-models for various time intervals which can lead to considerable computational expenses. Furthermore, as discussed in section 3.2, PINNs’ performance is limited to a specific realization of a PDE system and they lack the necessary flexibility for rapid adaptation to relevant tasks (e.g., different boundary HTCs or variation in fiber volume fraction in composites manufacturing). In order to address this computational bottleneck and improve the generalization and adaptability of PINNs, the proposed SMT method is employed and evaluated. Specifically, 20 relevant tasks (support set) in the curing of 1D composite parts are considered for training by varying the top- and bottom-side HTCs and fixing the rest of the process setting variables. Each training task is obtained by randomly sampling the top and bottom HTCs from the \textit{training distribution}, [40, 120] \(\frac{W}{m^{2}k}\) (Figure 5). Next, SMT’s meta-learners are trained throughout the decomposed temporal domain (as elaborated in section 3.3). Once trained, the learned parameters are used as the optimal initial state for fast and data-efficient adaptation to new curing processes with different boundary specifications (i.e., HTCs).

\textbf{\textit{Remark 1 – Meta-learners’ initialization}: }it was noticed that when training the first meta-learner (for the time interval 1) from scratch, the PDE and ODE loss components generate large error gradients which results in the exploding gradient effect and leads to infinite loss value. To avoid this, we switched off the PDE and ODE loss terms and trained on the initial and boundary points for 1000 epochs to ensure that the model locates in a more stable region in the optimization space and then switched on physical loss terms.

\textbf{\textit{Remark 2 – SMT training}}: it was realized that the learning rate of the SMT’s inner loop plays a significant role in the convergence of the meta-learners, especially for regions with stiff solutions. For all time intervals, the training started with an initial learning rate of 1$\times$10\textsuperscript{-5} and then using a stepwise annealing strategy, the learning rate is reduced by the order of 10 each time improvement was not observed over a specified number of epochs.

To evaluate SMT, its generalization performance is compared to that of MTL and TL approaches described in [37] and [24], respectively. For MTL, three tasks with three different sets of HTC values ([top HTC-bottom HTC]) are selected to be learned ([60-20]\textit{ }\(\frac{W}{m^{2}k}\), [120-70]\textit{ }\(\frac{W}{m^{2}k}\), [80-40]\textit{ }\(\frac{W}{m^{2}k}\)). To train MTL, a network with 6 neurons in the output layer (each task has two neurons for predicting the temperature and DoC) is utilized. The idea is that training on relevant tasks using a single network can encourage learning a hidden state with a more general representation of the solution space and thus, facilitates training the network on other similar tasks in a faster and more efficient way. For training a new task (e.g., a new set of HTC values), only the output layer is reduced to a two-neuron layer (i.e., initialized with the weights of one of the trained tasks) and the rest of the network (hidden layers) remains frozen. The network is then fine-tuned with the loss components of the new task. For TL, the source network is trained on a curing process with top and bottom HTCs set to 120 \(\frac{W}{m^{2}k}\) and 70 \(\frac{W}{m^{2}k}\), respectively, and then is used to initialize the target network for learning a text curing processes. Table 3 summarizes the predictive performance (relative \(\mathcal{L}^{2}\) error) of the above models against a test task with a symmetric (identical top and bottom values) HTC of 50 \(\frac{W}{m^{2}k}\). The models are fine-tuned for 1, 100 and 1000 iterations using 200 collocation, 40 boundary and 20 initial points per time interval. 

It is worth noting that the training datasets used for this evaluation are significantly smaller in size and the number of training epochs is much less compared to the standard training procedures commonly used in the literature for training PINNs. This is done in order to evaluate the performance of the models for rapid and efficient adaptation to new tasks with a few gradient steps and very limited training data. As the number of training points and training iterations increases, the performance of all models is expected to enhance accordingly. The results show that SMT requires as few as one epoch to efficiently adapt to the new task. This is not achieved with TL and MTL as they need a longer training period to yield the same performance. This is due to the fact that the meta-learners in SMT are specifically trained to enable fast adaptations to novel scenarios with only a few gradient iterations (in this study, we chose one epoch of fine-tuning for optimizing the meta-learner’s inner loop). As the number of epochs increases, TL and MTL begin to perform better and obtain similar performance to SMT. This is also expected as TL-based fine-tuning improves with longer training periods and more iterations \cite{yosinski2014how}. Table 3 also compares the model’s training computational cost in terms of the number of epochs per second. MTL is the slowest as its loss function comprises of 3 PDE loss terms (each per task, as opposed to 1 PDE term for conventional PINN). SMT’s training is slower than that of TL as it requires computing gradients for both inner and outer loops for updating its parameters. TL, however, does not add any additional computational costs to the conventional PINN setting as it only fine-tunes the source parameters. Finally, Table 3 also shows how many epochs TL and MTL need on average to achieve the similar Relative $L^2$ error achieved by SMT after 1 epoch. Clearly, SMT is the dominant method for fast and efficient adaptation across different tasks in PINNs.

\begin{table}[h]
\centering
\caption{Comparing predictive performance of TL, MTL and SMT against a test task with symmetric HTC of $50 \frac{w}{m^2 k}$.}
\begin{tabular}{cccccc}
\hline 
\multirow{2}{*}{ PINN model } & \multicolumn{3}{c}{ Temperature prediction relative $\mathcal{L}^2$ error } & \multirow{2}{*}{Training time} & \multirow{2}{*}{Adaptation epochs}\\
\cmidrule(rl){2-4}
& 1 epoch & 100 epochs & 1000 epochs & & \\
\hline TL [24] & $1.2 \times 10^{-2}$ & $4.3 \times 10^{-3}$ & $1.9 \times 10^{-3}$ & \textbf{95} & 115 \\
MTL [37] & $1.4 \times 10^{-2}$ & $3.8 \times 10^{-3}$ & $1.8 \times 10^{-3}$ & 31 & 108 \\
\textbf{SMT (ours)} & \boldmath{$3.2 \times 10^{-3}$} & \boldmath{$2.1 \times 10^{-3}$} & \boldmath{$1.5 \times 10^{-3}$} & 44 & \textbf{1} \\
\hline
\end{tabular}
\end{table}

The training process of SMT can be further investigated. Due to MTL’s relatively high computational cost during the training phase and its comparable predictive performance to TL, in the following analyses, only SMT and TL are compared and evaluated as the most viable models. In particular, as indicated in Figure 5, two in-distribution (symmetric 50\(\frac{W}{m^{2}k}\) and symmetric 60\(\frac{W}{m^{2}k}\)) and one out-of-distribution (symmetric 40\(\frac{W}{m^{2}k}\)) tasks are selected for evaluation. Figure 6 shows the models’ performance during the first 100 epochs of adaptation to the test tasks. Specifically, for each time interval, the corresponding meta-learner/source model is fine-tuned for 100 epochs (12 intervals, hence 1200 epochs in total) and the test performance on temperature prediction during the training is measured. The relative \(\mathcal{L}^{2}\) error and maximum absolute error of SMT and TL under each task are shown in the background, which are overlayed by the average performance indicated by thick red and black lines. In all time intervals, SMT outperforms TL significantly within the first few epochs. Specifically, in the ``stiff" region of the curing process with sharp transition (minutes 120 to 180 in Figure 5 corresponds to epochs 500 to 800 in Figure 6), SMT was able to keep the temperature’s maximum absolute error of in-distribution tasks below 5$^{\circ}$C. Away from the first epochs, although SMT continues to improve the prediction performance, it can be seen that TL’s error reduction occurs at a faster rate. This is expected as meta-learning models are designed to yield accurate predictions with only a few gradient steps, while TL models usually achieve this in the long run with more epochs. Regardless, because of the early advantage, SMT maintains its upper hand against TL even in longer training periods.

\begin{figure}[!htbp]
\centering
\includegraphics[width=13cm]{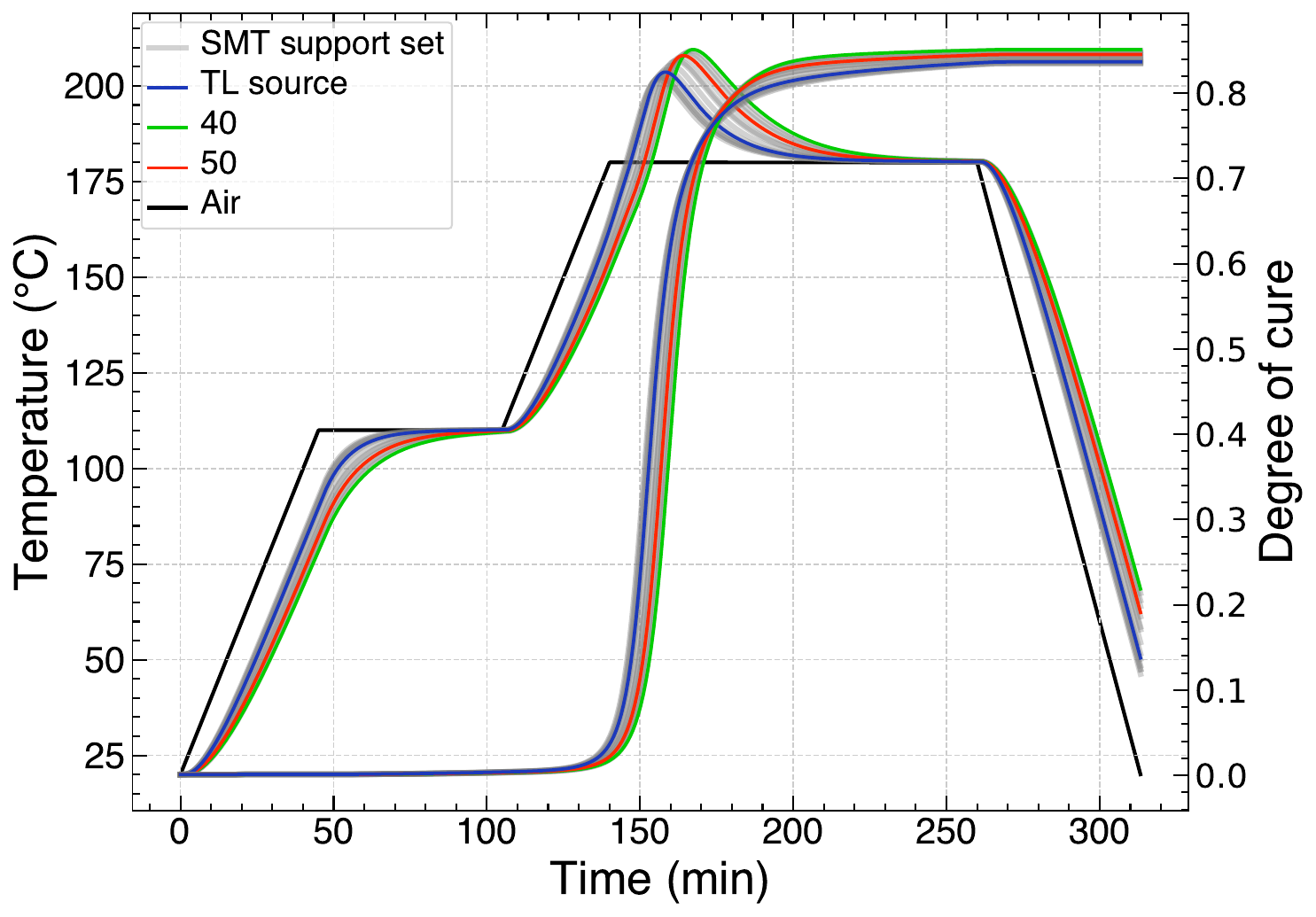}
\caption{Temperature and DoC profiles of SMT support set (training tasks), TL source task, in-distribution (50\(\frac{W}{m^{2}k}\)) and out-of-distribution (40\(\frac{W}{m^{2}k}\)) test tasks. The profiles indicate the behaviour of the material at its middle section.}
\label{fig:temperature_and_doc_profiles_smt}
\end{figure}

\begin{figure}[!htbp]
\centering
\includegraphics[width=16cm]{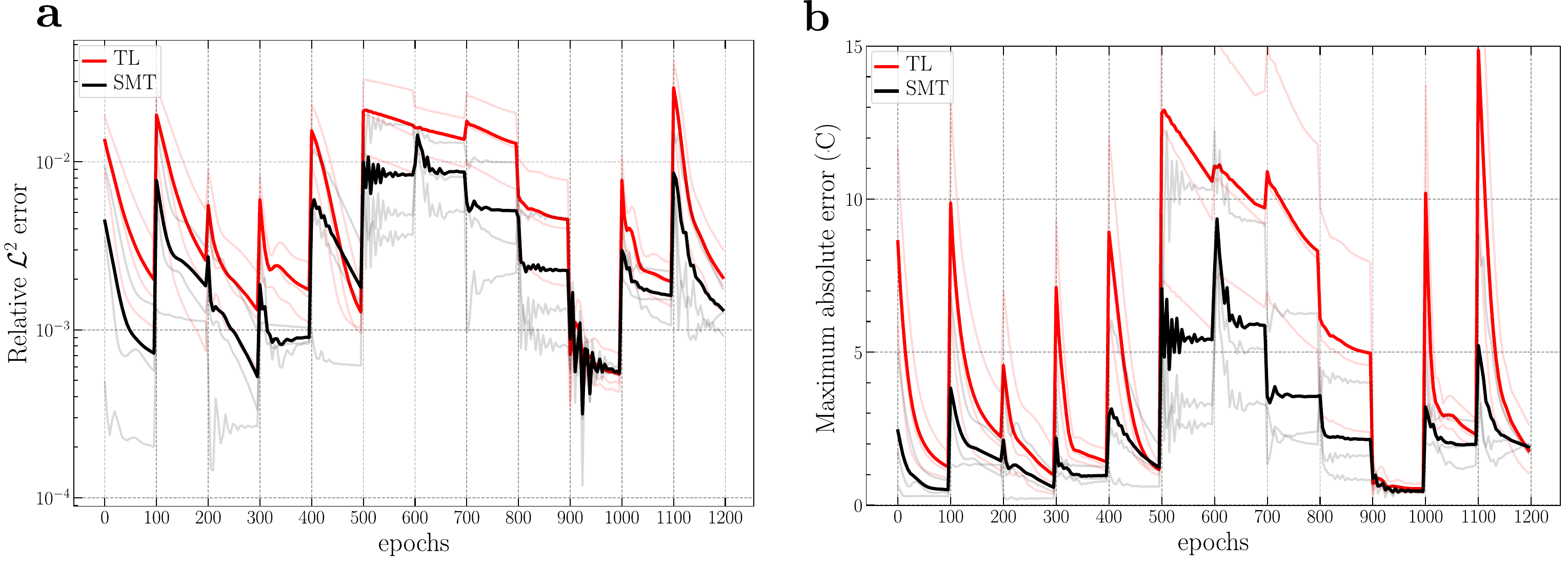}
\caption{Comparison of SMT and TL performance in temperature prediction. Three test scenarios, namely two in-distribution tasks (symmetric 50 and 60) and one out-of-distribution task (symmetric 40) are used and relative $L_2$ error (a) and maximum absolute error (b) values are recorded. The individual performances for each case are demonstrated by shades in the background which are overlaid by thick red and black lines representing the average performances. Each time interval is trained for 100 epochs (hence 1200 epochs in total) }
\label{fig:comparison_smt_and_tl_performance}
\end{figure}

Figure 7 compares TL and SMT's temperature and DoC prediction performance at the first 2 epochs of fine-tuning for the test tasks. FE simulation results are presented as the reference. SMT was able to better capture the part's thermochemical behaviour, especially at the stiff regions as well as the part's maximum temperature (exotherm). While going from epoch 1 to epoch 2 does not change the TL curves much, SMT exhibits a considerable improvement in its prediction performance. This is again, a clear indication of SMT's success in fast and efficient adaptation. Moreover, SMT shows a more robust behaviour against the out-of-distribution task (Figure 7.c and f) and does not deviate from the true response as much as TL did. This also highlights a shortfall of TL in knowledge transfer as it performs poorly if the difference between the source and target distributions is considerable. In other words, in such scenarios, TL might require much longer training periods in order to fully adapt to the new task. It also should be pointed out that a few discontinuities (mainly around the stiff region) are evident in SMT’s predictions. This is because, in contrast to TL which leverages fully converged source weights from similar tasks, SMT uses meta-learners trained on a variety of tasks, each with a unique initial condition (as it depends on the predictions in the previous time interval). This requires a few more epochs for the SMT to reduce the initial condition loss appropriately and yield a more continuous prediction. Regardless, the overall performance of SMT clearly dominates that of TL.

\begin{figure}[!htbp]
\centering
\includegraphics[width=16cm]{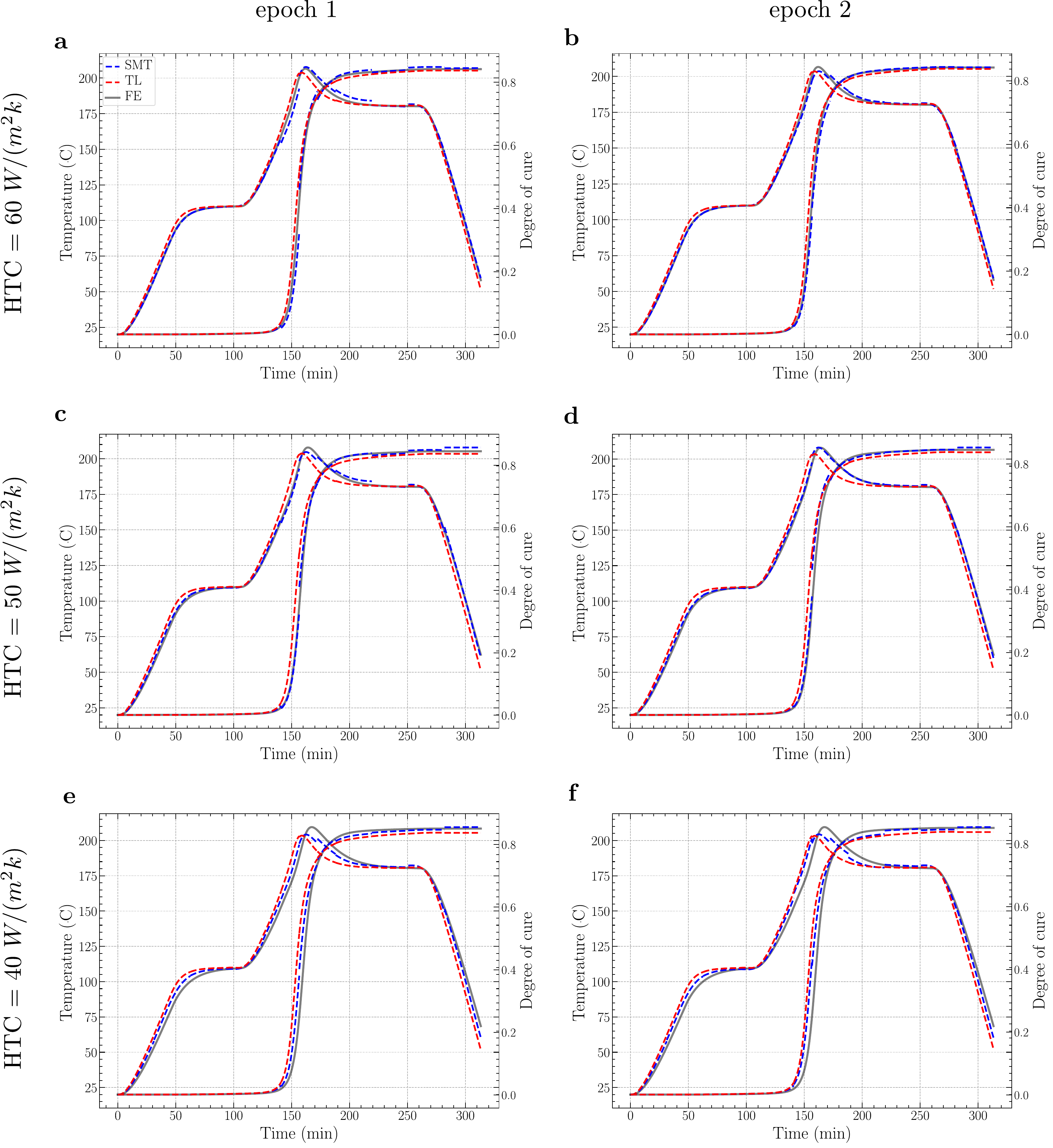}
\caption{Temperature and DoC prediction performance of SMT and TL after 1 (a, c, e) and 2 (b, d, f) epochs of finetuning against the test tasks. FE simulation result is presented for reference.}
\label{fig:temperature_and_doc_prediction_performance}
\end{figure}

\section{Conclusion}

This work presented a new sequential meta-transfer-learning approach that can concurrently address the PINN’s frequently faced poor performance in solving ‘stiff’ problems with long temporal domains, as well as the PINN’s slow and costly adaptations/generalization to new tasks (e.g., when the system parameters or boundary conditions change). This unified framework breaks down the input domain into smaller time segments and hence easier PDE problems. It then trains the PINN model sequentially over all time intervals using a set of subnetworks. The learning framework, on the other hand, leverages the capabilities of meta-transfer learning for fast and efficient adaptations to new tasks (e.g., systems with different boundary conditions). Specifically, at each time interval, instead of training a conventional PINN, the framework learns a meta-learner which produces optimal initial weights suitable for fast adaptation tasks with only a few gradient steps. The presented hybrid and adaptive learning method was evaluated in an advanced composites autoclave processing case study, and it exhibited a dominant performance in comparison to other approaches in the literature. Namely, in comparison to TL and MTL, the proposed SMT method expedites the task adaptation of PINN models by a factor of 100. Finally, one way to improve the accuracy of the proposed method is to incorporate \textit{hard }constraints to satisfy the initial conditions. In sequential learning, the initial condition of each time segment is determined by the approximation of the sub-network trained in the previous time interval. Also, the training of the PINN model via minimizing the loss function can still result in some residual error in the initial condition. This can cause error propagation over the time intervals and thus yield considerable deviation from the system’s true behaviour. Utilizing hard constraints can alleviate such effects by strictly enforcing the initial conditions (no residual error) for the training of each sub-network.

\section{Acknowledgments}

This study was financially supported by the New Frontiers in Research Fund – Exploration stream (award number: NFRFE-2019-01440).

\bibliographystyle{unsrt}  
\bibliography{bibliography-bibtex}

\end{document}